\documentclass[12pt,a4paper]{article}
\usepackage[utf8]{inputenc}
\usepackage{amsmath}
\usepackage{graphicx}
\usepackage[margin=1in]{geometry}
\usepackage{caption}
\usepackage{svg}
\usepackage{subcaption}
\usepackage{hyperref}
\usepackage{authblk}
\usepackage{booktabs}

\title{PReLU: Yet Another Single-Layer Solution to the XOR Problem}
\author{Rafael C. Pinto$^1$, Anderson R. Tavares$^2$ \\
    \small \texttt{rafael.pinto@canoas.ifrs.edu.br$^1$, artavares@inf.ufrgs.br$^2$}\\
    \small Federal Institute of Rio Grande do Sul (IFRS)$^1$ \\
    \small Canoas, Brazil$^1$ \\
    \small Institute of Informatics, Federal University of Rio Grande do Sul (UFRGS)$^2$ \\
    \small Porto Alegre, Brazil$^2$ \\
}
\date{\today}

\begin{document}

\maketitle

\begin{abstract}
This paper demonstrates that a single-layer neural network using Parametric Rectified Linear Unit (PReLU) activation can solve the XOR problem, a simple fact that has been overlooked so far. We compare this solution to the multi-layer perceptron (MLP) and the Growing Cosine Unit (GCU) activation function and explain why PReLU enables this capability. Our results show that the single-layer PReLU network can achieve 100\% success rate in a wider range of learning rates while using only three learnable parameters.
\end{abstract}

\section{Introduction}
The XOR problem has traditionally been used to illustrate the limitations of single-layer networks since Minsky and Papert's seminal work \cite{Minsky1969}, which even contributed to the first AI Winter \cite{toosi2021brief}. It has traditionally required at least one hidden layer to solve, making it a litmus test for network complexity. Trivially, any function, no matter how complex, can be learned in a single layer by just using itself as the activation function, and that says nothing about its general applicability and usefulness. Here, however, we reveal this ability in a simple, general and well-established activation function. This study demonstrates how using the Parametric Rectified Linear Unit (PReLU) activation \cite{He2015} overcomes these limitations, effectively solving the XOR problem without additional layers. This ability has significant implications for neural network design and efficiency, potentially leading to simpler architectures for complex problems.

On another front, recent advancements in neuroscience have revealed that individual human neocortical pyramidal neurons can learn to compute the XOR function \cite{Gidon2020}. This discovery has inspired new artificial neuron models and activation functions that aim to bridge the gap between biological and artificial neurons \cite{Noel2021b}. Albeit not producing the same activation curves as the ones found in biological neurons, the PReLU activation matches their representational power, at least regarding the XOR function.

\section{Background}

Activation functions play a crucial role in neural networks by introducing non-linearity, allowing networks to learn complex patterns. Recent surveys have cataloged hundreds of activation functions \cite{Kunc2024}, highlighting the diversity and ongoing research in this area. 

Of special interest to us is the Growing Cosine Unit (GCU), a recently proposed oscillatory activation function defined as $f(x) = x\cos(x)$ \cite{Noel2021}. GCU has shown promising results in various benchmarks and can solve the XOR problem with a single neuron, and thus will serve as a comparison to PReLU results. Other oscillating activation functions capable of solving XOR were presented in \cite{Noel2021b}, but were left out since GCU was the best overall performer in that work.

\subsection{Parametric ReLU (PReLU)}
PReLU, introduced by He et al. \cite{He2015}, is defined as:

\[
f(x) = 
\begin{cases} 
x & \text{if } x \geq 0 \\
ax & \text{if } x < 0
\end{cases}
\]
 


where $a$ is a learnable parameter. This allows flexibility, enabling a single-layer solution to the XOR problem. Besides, PReLU generalizes various other activation functions, depending on the parameter $a$:

\begin{itemize}
\item When $a = 0$, PReLU becomes the standard ReLU function.
\item When $a = 1$, PReLU becomes the identity function. It might be useful for accumulators or copying values between layers, as well as layer pruning \cite{dror2021layer}.
\item When $0 < a < 1$, PReLU becomes the LeakyReLU function \cite{maas2013rectifier}.
\item When $a < 0$, PReLU allows for non-monotonic behavior, which is crucial for solving the XOR problem in a single layer.
\item When $a = -1$, PReLU becomes the absolute value function, as we show below.
\end{itemize}

PReLU can also be written as $f(x) = max(0, x) + a \cdot min(0, x)$. When  $a=-1$, PReLU becomes $f(x) = \max(0, x) + (-1) \cdot \min(0, x) = \max(0, x) - \min(0, x)$. Now, let's consider two cases:
\begin{itemize}
\item When $x \geq 0$, $f(x) = \max(0, x) - \min(0, x) = x - 0 = x = |x|$
\item When $x < 0$, $f(x) = \max(0, x) - \min(0, x) = 0 - x = -x = |x|$
\end{itemize}

\begin{figure}[ht]
\centering
\begin{minipage}{0.45\textwidth}
    \centering
    \includegraphics[width=\textwidth]{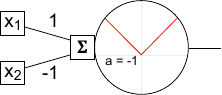}
    \caption{PReLU solution to the XOR problem with inputs in $\{0,1\}$. Connection weights are interchangeable. Halve connection weights for inputs in $\{-1,1\}$.}
    \label{fig:xor-prelu}
\end{minipage}
\hspace{0.05\textwidth}%
\begin{minipage}{0.45\textwidth}
    \centering
    \vspace{0.01\textwidth}%
    \begin{tabular}{|c|c|c|c|}
    \hline
    $x_1$ & $x_2$ & $x_1 \oplus x_2$ & $|x_1 - x_2|$ \\
    \hline
    0 & 0 & 0 & 0 \\
    0 & 1 & 1 & 1 \\
    1 & 0 & 1 & 1 \\
    1 & 1 & 0 & 0 \\
    \hline
    \end{tabular}
    \captionof{table}{Truth table for XOR operation and $|x_1 - x_2|$, which is equivalent to the PReLU activation for $a = -1$.}
    \label{tab:xor_truth_table}
\end{minipage}%
\end{figure}

Therefore, for all real numbers $x$, PReLU with $a=-1$ is equivalent to the absolute value function $|x|$. This is further illustrated in figure \ref{fig:xor-prelu} and table \ref{tab:xor_truth_table}, which also show how to solve the XOR problem using the $abs$ function. Also relevant: $|x| = ReLU(x) + ReLU(-x)$ and, more generally, $PReLU(x) = a\cdot x + (1-x)\cdot ReLU(x)$, which shows that a single PReLU neuron is equivalent to a sum of 2 monotonic neurons (both non-linear in the absolute function case), potentially reducing network size.

\section{Methodology}

Aside from the theoretical equivalences previously shown, we wanted to verify PReLU's performance in the XOR problem empirically. We implemented \footnote{\url{https://colab.research.google.com/drive/1-zwn-IX4iC3LVQ77u-FENu5lQHnFMhzv?usp=sharing}} and compared three networks in PyTorch \cite{paszke2017automatic} using the following configurations obtained through hyperparameter optimization \cite{optuna_2019}:
\begin{itemize}
\item Single-Layer PReLU: 2 input neurons, 1 output neuron with PReLU (no bias). Three learnable parameters in total. Inputs in the range $\{-1,1\}$.
\item Single-Layer GCU: 2 input neurons, 1 output neuron with GCU. Three learnable parameters in total. Inputs in the range $\{0,1\}$.
\item MLP: 2 input neurons, 2 hidden neurons with Tanh activation \cite{Lecun1989} (Lecun's variant, which learned better and faster than ReLU in this experiment), 1 output neuron (no output bias). Eight learnable parameters in total. Inputs in the range $\{-1,1\}$.
\end{itemize}

All networks were trained for 300 epochs using the Adam optimizer and Mean Squared Error (MSE) loss. Biases and $a$ are initialized at $0$. We conducted 100 experiments using different random initializations for each network. Networks' performance across different learning rates and input ranges was also analyzed.

\section{Results}

\begin{figure}[htb]
    \centering
    \begin{minipage}[t]{0.48\textwidth}
        \centering
        \includegraphics[width=\textwidth]{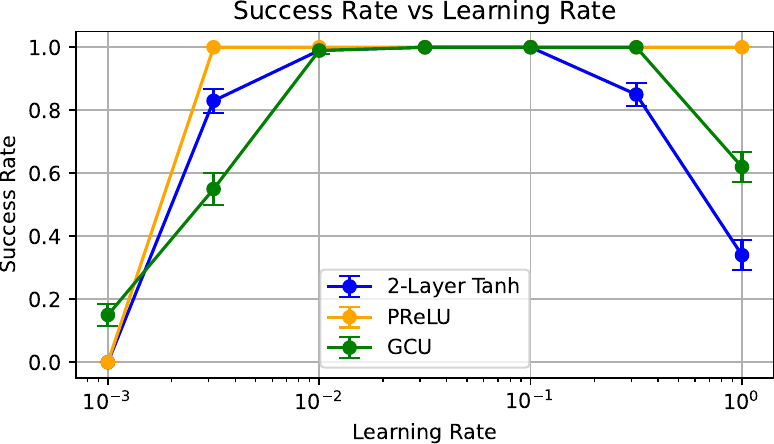}
        \caption{Success rate vs learning rate (100 trials, 300 epochs). PReLU shows the widest range of acceptable learning rates for the Adam optimizer in the XOR problem. In a separate experiment, PReLU with a bias shows similar behavior to the other models when close to learning rate 1.}
        \label{fig:accuracy_lr_plot}
    \end{minipage}
    \hfill
    \begin{minipage}[t]{0.48\textwidth}
        \centering
        \includegraphics[width=\textwidth,height=130px]{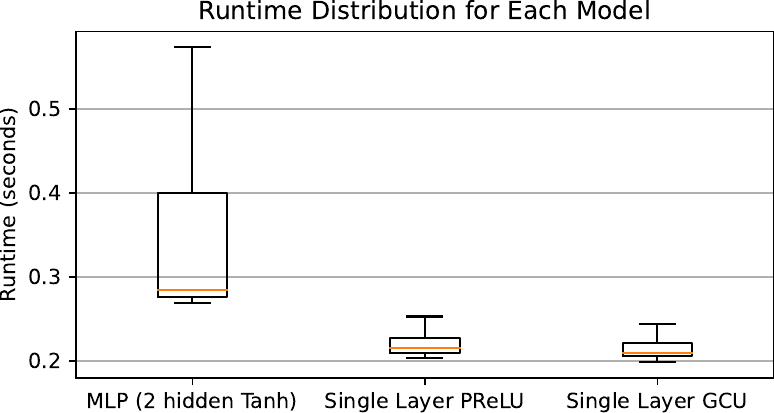}
        \caption{Runtime distribution for each model for a single trial (300 epochs). PReLU and GCU (3 learnable parameters each) are significantly faster than the MLP solution (8 learnable parameters) and have similar runtime.}
        \label{fig:runtime}
    \end{minipage}
\end{figure}

\begin{figure}[htb]
    \centering
    \begin{subfigure}[t]{0.48\textwidth}
        \centering
        \includegraphics[width=\textwidth]{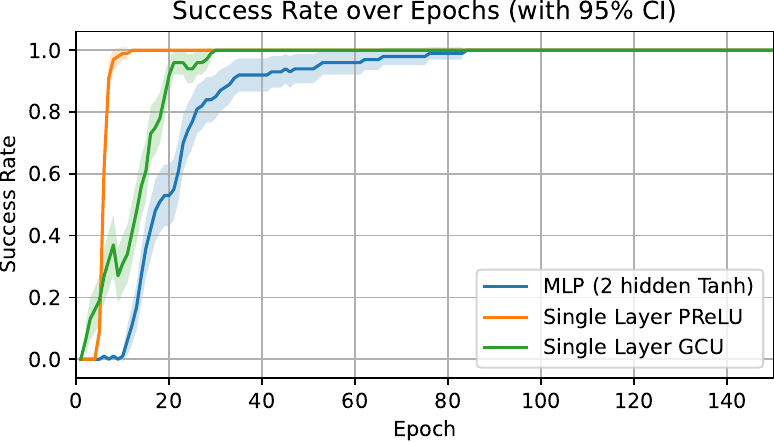}
        \caption{Success rate: PReLU converges in less than 20 epochs, while GCU goes through some detours before solving the problem close to 50 epochs.}
        \label{fig:accuracy_epochs_plot}
    \end{subfigure}
    \hfill
    \begin{subfigure}[t]{0.48\textwidth}
        \centering
        \includegraphics[width=\textwidth]{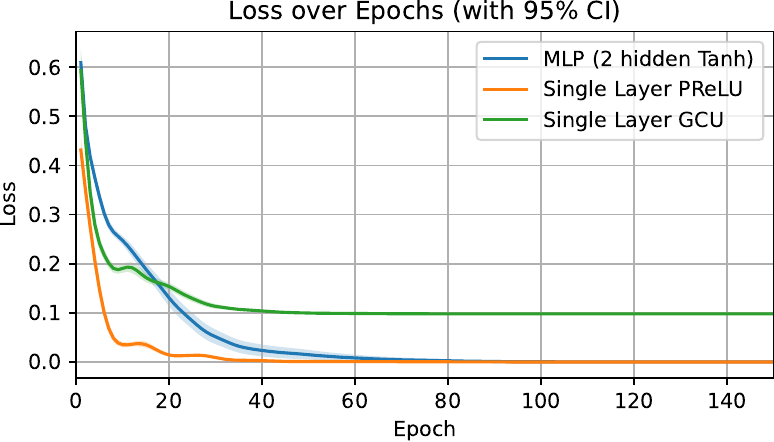}
        \caption{Mean Squared Error: GCU is the only model unable to reach zero error, likely due to its locally bounded nature around the origin.}
        \label{fig:loss_epochs_plot}
    \end{subfigure}
    \caption{Comparison of success rate and MSE over 150 epochs and 100 trials.}
    \label{fig:combined_plots}
\end{figure}

\begin{figure}[htb]
\centering
\includegraphics[width=1\textwidth]{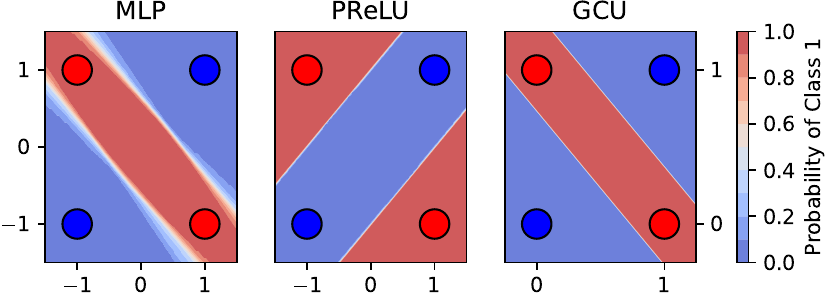}
\caption{Average decision boundaries over 100 trials. PReLU and GCU show the least variance in learned solutions, as expected due to their lower number of learnable parameters. PReLU obtained the widest margins between classes, indicating more robustness to noise and better generalization capability. Note that $\{0,1\}$ inputs were used for GCU instead of $\{-1,1\}$, as it gave the best results for this model.}
\label{fig:decision_boundaries}
\end{figure}

\begin{figure}[!htb]
\centering
\begin{minipage}{0.33\textwidth}
\centering
\includegraphics[width=\textwidth]{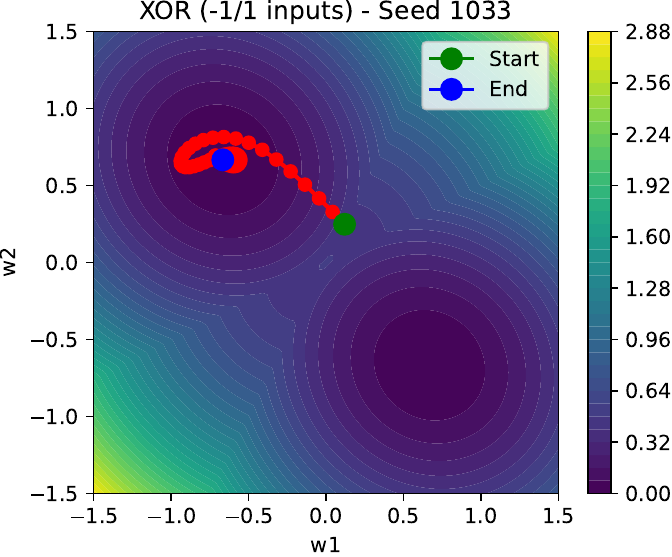}
\end{minipage}%
\begin{minipage}{0.33\textwidth}
\centering
\includegraphics[width=\textwidth]{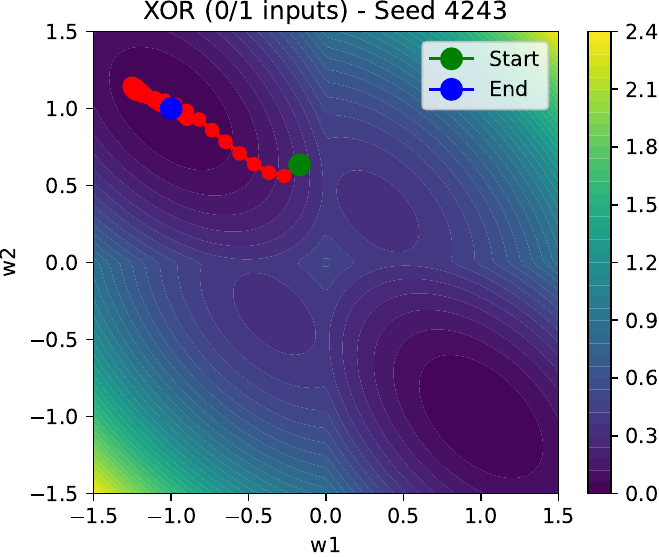}
\end{minipage}%
\begin{minipage}{0.33\textwidth}
\centering
\includegraphics[width=\textwidth]{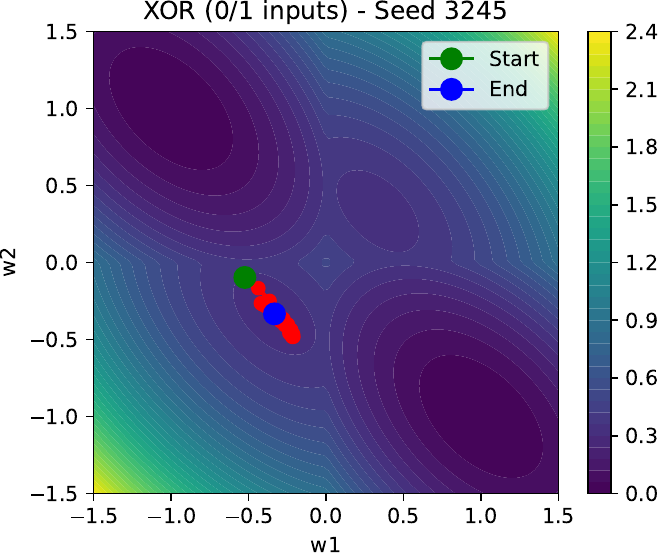}
\end{minipage}
\caption{Learning behavior for the PReLU activation (with $a = -1$) with different input ranges. \textbf{Left}: Input range ${-1,1}$. There are two global minima and no local minima close to the origin, thus, the problem can always be solved. \textbf{Middle}: Input range ${0,1}$. Two of the four quadrants have local minima corresponding to the AND function. When the initial weights start in the 2nd or 4th quadrants, however, Adam finds the global minima. \textbf{Right}: Starting in the 1st or 3rd quadrants results in failure.}
\label{fig:three_images}
\end{figure}

PReLU achieved a 100\% success rate in solving XOR across a wide range of learning rates (Figure \ref{fig:accuracy_lr_plot}), while converging faster than both GCU and MLP, reaching 100\% accuracy in under 20 epochs (Figure \ref{fig:accuracy_epochs_plot}). It also achieved zero loss, while GCU could not, probably due to its locally bounded nature around the origin (Figure \ref{fig:loss_epochs_plot}). Both PReLU and GCU showed less variance in learned decision boundaries compared to MLP, which is expected due to their lower number of learnable parameters (Figure \ref{fig:decision_boundaries}), and both were significantly faster to train than MLP also due to smaller network size (Figure \ref{fig:runtime}).

The learning behavior of PReLU (Figure \ref{fig:three_images}) reveals that the input range and initial weights play crucial roles in convergence. For inputs in $\{-1,1\}$, PReLU always converges to a global minimum. However, for inputs in $\{0,1\}$, the convergence depends on the initial parameter quadrant, with some quadrants leading to local minima corresponding to the AND function.

\section{Conclusion}

This study demonstrates that by using PReLU, a single-layer network gains the flexibility to implement non-monotonic functions when necessary, enabling it to solve the XOR problem efficiently, obtaining a 100\% success rate in a wide range of learning rates with fewer parameters. This insight may be valuable in designing efficient neural architectures for problems traditionally thought to require multiple layers, as also concluded in  \cite{Noel2021}. This aligns with recent discoveries in neuroscience about the computational capabilities of individual neurons \cite{Gidon2020} and supports the development of more biologically inspired artificial neurons \cite{Noel2021b}.

Future work could explore the implications of these findings for more complex problems and larger networks, aiming at potentially smaller training times and less energy consumption. Additionally, investigating other activation functions that enable single-layer solutions to traditionally multi-layer problems without added complexity could lead to new paradigms in neural network design.

\bibliographystyle{abbrv}
\bibliography{export}

\end{document}